%% file: main.tex
\documentclass{article}

\usepackage[final]{corl_2026} 
\usepackage[shortlabels,inline]{enumitem}
\usepackage{amsmath,amssymb}
\usepackage{graphicx}
\usepackage{booktabs}
\usepackage{caption}
\graphicspath{{figs/}}
\usepackage{fontawesome5}
\newcommand{\shield}{\text{\scalebox{0.6}{\faShield*}}}
\newcommand{\fallback}{\pi^{\shield}}
\newcommand{\monitor}{\Delta^{\shield}}
\newcommand{\safeset}{\Omega^{\shield}}
\newcommand{\task}{\pi^{\text{task}}}
\newcommand{\conf}{\beta^{\text{task}}}
\usepackage{amsthm}
\theoremstyle{definition}
\newtheorem{definition}{Definition}
\theoremstyle{plain}

\newtheorem{hypothesis}{Hypothesis}
\theoremstyle{remark}

\title{Confidence-Modulated Safety Filtering for End-to-End Policies}

\author{
  Javier Borquez \hspace{2.5em} Jaime F. Fisac\\
  \mdseries Department of Electrical and Computer Engineering, Princeton University\\
  \mdseries\texttt{\{javier.borquez, jfisac\}@princeton.edu}
}

\begin{document}
\maketitle


\begin{abstract}
    End-to-end policies, from task-specific controllers to robot foundation models, learn complex
    behavior from raw observations but are not guaranteed to remain safe, so they are often wrapped
    in a safety filter.
    The resulting system operates under two domains: the filter's operational design domain (ODD),
    where its assumptions hold and safety is certified, and the policy's competence domain, where it
    is reliable.
    Filters typically fix their conservativeness regardless of the policy's competence, trading
    performance inside the ODD against margin beyond it.
    We ask whether the policy's confidence should set that level, and propose a filter-agnostic
    pipeline that modulates it at runtime: inside the ODD, where every setting is safe, confidence
    buys performance; beyond it, low confidence adds margin, which helps when it flags where the
    filter's model fails.
    Preliminary F1TENTH simulations, where this holds, show the modulated filter combining the
    strengths of permissive and conservative filters.
\end{abstract}



\input{sections/01_intro}
%
%
%
\input{sections/04_method}
%
\input{sections/05_experiments}

%
\input{sections/06_conclusion}


\clearpage


\bibliography{references}  

\end{document}

%% file: sections/01_intro.tex

\section{Introduction}
\label{sec:intro}

End-to-end policies that map observations directly to actions have become a practical route to
capable autonomous decision-making, from task-specific
controllers~\citep{bojarski2016end,kaufmann2023champion,geles2024demonstrating} to generalist robot
foundation models~\citep{zitkovich2023rt2,kim2024openvla,black2025pi0}.
Trained on enough data, they often perform well and, much of the time, act safely.
But not always: their competence is tied to the domain of the inputs they were trained on, and away
from it their behavior is unspecified.
A racing policy that drives a familiar track flawlessly may steer into a wall once the lighting
changes or the tires lose grip~\citep{filos2020can,dehaan2019causal}.
Worse, such failures arrive without warning: the policy issues an action exactly as it always does,
and nothing in its action signals that it has left the regime in which its behavior was
learned~\citep{hecker2018failure,agia2024unpacking}.
A standard remedy for safety-critical deployment is to wrap the policy in a \emph{safety filter}
that monitors its proposed actions and overrides them when they threaten to drive the system into
failure~\citep{hsu2024safety,borquez2024safety,ames2019control}.

A safety filter, however, is only as good as the assumptions behind it.
Its certificate rests on a model of the world, such as the system dynamics, a map of free space, a
bound on disturbances, and holds only within the operational design domain (ODD) where those
assumptions are met.
Inside the ODD, the filter keeps the system safe regardless of the policy it guards; beyond it, the
filter keeps acting, but nothing is guaranteed~\citep{hsu2024safety,sae2021j3016}.
The learned policy has its own competence domain, the conditions under which its training
makes its behavior reliable.
The two domains do not need to coincide, yet standard safety-filtered pipelines treat the policy as an
arbitrary black box and ignore whether it is operating within its competence
domain~\citep{wabersich2023data,wabersich2021predictive}.

Blind to the policy's competence, standard filters commit to a single level of conservativeness, set
by design parameters such as a barrier decay rate or a safety-value
threshold~\citep{bansal2017hamilton,alan2022safe}.
Inside the ODD every setting is safe, but a conservative filter overrides a competent policy early
and often, eroding its performance, while a permissive one lets an unreliable policy reach the
boundary of the safe set, where abrupt fallback interventions are
costly~\citep{hu2022sharp,hu2024active}.
Beyond the ODD, losing the certificate does not make failure inevitable: the filter becomes an
imperfect one, and its conservativeness sets a margin that may or may not translate into safety.
On a race track, a wide margin may save a car whose filter assumes dry asphalt once the track turns
wet, yet it also brakes the car on straights the policy has driven thousands of times.
Whatever value is chosen is fixed and spent uniformly, whether or not the policy has proven reliable
there.

\begin{figure}[t]
    \centering
    \includegraphics[width=0.94\linewidth]{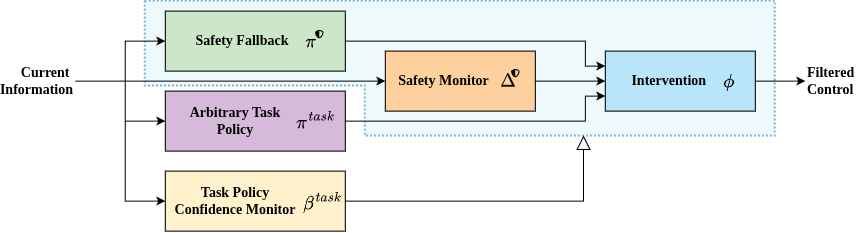}
    \caption{Confidence-modulated safety filtering. A confidence monitor
        $\beta^{\text{task}}$ estimates how much the task policy $\pi^{\text{task}}$ can be trusted
        on the current information and modulates the conservativeness of the safety filter
        (fallback $\pi^{\shield}$, monitor $\Delta^{\shield}$, intervention $\phi$). Adapted
        from~\citet{hsu2024safety}.}
    \label{fig:pipeline}
\end{figure}

A fixed conservativeness has been questioned before, with adaptation driven by
performance~\citep{parwana2025rate,kim2025learning,kim2025adapt}, by the filter's own prediction
errors~\citep{huriot2026safe}, or by the policy's own uncertainty, which can tighten a CBF
filter~\citep{yu2025uncertainty,li2026risk} or trigger a switch to a fallback when inputs look
novel~\citep{richter2017safe}.
Building on this line of work, we ask whether the task policy's confidence should set a safety filter's
conservativeness: a confidence monitor $\beta^{\text{task}}$ estimates how much the task policy can
be trusted on the current input and sets the filter's conservativeness (Fig.~\ref{fig:pipeline}).
When confidence is high, the policy is likely reliable and the filter steps back; when it is low, the
policy's behavior is unpredictable and conservativeness costs little, provided that low confidence
signals risk.
Inside the ODD, safety is untouched, since every setting is certified, and modulation only buys
performance; beyond it, modulation adds margin where the policy is less trustworthy.
We give a filter-agnostic formulation, for any confidence measure and any filter whose
conservativeness can be graded, and use the two domains to make explicit what modulation buys
inside and beyond the ODD, and under which conditions.

%% file: sections/04_method.tex

\section{Confidence-Modulated Safety Filtering}
\label{sec:method}


\paragraph{Safety filters.}
Following~\citet{hsu2024safety}, we consider a system with state
$x_t \in \mathcal{X} \subseteq \mathbb{R}^{n_x}$ and control input
$u_t \in \mathcal{U} \subseteq \mathbb{R}^{n_u}$, evolving as $x_{t+1} = f(x_t, u_t, d_t)$, where the
disturbance $d_t \in \mathcal{D}$ captures modeling error and exogenous effects.
The state is not observed directly: all knowledge available at time $t$ is encoded in an information
state $\eta_t \in \mathcal{H}$, which evolves as $\eta_{t+1} = f^\eta(\eta_t, u_t, d_t)$; below,
$\eta$ denotes a generic element of $\mathcal{H}$.
A failure set $\mathcal{F} \subseteq \mathcal{X}$ collects the failure states, and its
information-space counterpart $\mathcal{F}^\eta \subseteq \mathcal{H}$ the information states
consistent with some $x \in \mathcal{F}$.
Our nominal end-to-end policy is the task policy $\task : \mathcal{H} \to \mathcal{U}$.
A safety filter comprises a fallback policy $\fallback : \mathcal{H} \to \mathcal{U}$, a safety
monitor $\monitor : \mathcal{H} \times \mathcal{U} \to \mathbb{R}$, and an intervention scheme
$\phi : \mathcal{H} \times \mathcal{U} \to \mathcal{U}$.
The monitor is \emph{valid}~\citep[Def.~1]{hsu2024safety} if $\monitor(\eta, u) \geq 0$ implies
$\eta \notin \mathcal{F}^\eta$ and $f^\eta(\eta, u, d) \in \safeset$ for all $d \in \mathcal{D}$,
where $\safeset \subseteq \mathcal{H}$ contains the information states from which $\fallback$ keeps
the system out of $\mathcal{F}$ under every disturbance; the triple satisfies the \emph{filter
condition}~\citep[Def.~2]{hsu2024safety} if
$\monitor(\eta, \fallback(\eta)) \geq 0 \Rightarrow \monitor(\eta, \phi(\eta, u)) \geq 0$ for all
$u \in \mathcal{U}$.
We call $\eta$ \emph{certified} if $\monitor\big(\eta, \fallback(\eta)\big) \geq 0$ and, as is
implicit in~\citep[Thm.~1]{hsu2024safety}, assume that passing the check leads to certified states:
$\monitor(\eta, u) \geq 0 \Rightarrow f^\eta(\eta, u, d)$ is certified for all $d \in \mathcal{D}$.
Together, these guarantee all-time safety for every task policy from any certified
$\eta_0$~\citep[Thm.~1]{hsu2024safety}.

\paragraph{Operational design domain.}
We call the operational design domain (ODD) the set of assumptions under which $\monitor$ is valid:
the dynamics $f$, the disturbance set $\mathcal{D}$, a given perception stack, among others.
Beyond the ODD, $\monitor$ can potentially misrepresent the safety of the system.

\paragraph{Graded safety filters.}
We fix a base safety filter $(\fallback, \monitor, \phi)$ and assume its monitor is
\emph{margin-valued}: $\monitor(\eta, u)$ measures the margin to the unsafe set the system keeps
after applying $u$.

\begin{definition}[Conservativeness]
    \label{def:conservativeness}
    A safety filter $(\fallback_a, \monitor_a, \phi_a)$ is at least as conservative as
    $(\fallback_b, \monitor_b, \phi_b)$, written $a \succeq b$, if
    $\monitor\big(\eta, \phi_a(\eta, u)\big) \geq \monitor\big(\eta, \phi_b(\eta, u)\big)$
    for every certified $\eta$ and every $u \in \mathcal{U}$,
    where $\monitor$ is the base monitor.
\end{definition}
That is, whatever the task policy proposes, a more conservative filter never leaves less margin,
measured by the base monitor.
Each filter acts through its intervention scheme, which queries its own monitor and may apply its
own fallback, so changes to any of the three components are captured by the action it executes.

\begin{definition}[Graded safety filter family]
    \label{def:graded}
    A family of safety filters $\{(\fallback_\theta, \monitor_\theta, \phi_\theta)\}_{\theta \in
    \Theta}$, with $\Theta$ totally ordered, is \emph{graded} with respect to the base filter if
    \begin{enumerate*}[(a)]
        \item every $(\fallback, \monitor, \phi_\theta)$ satisfies the filter condition, and
        \item $\theta \geq \theta' \Rightarrow \theta \succeq \theta'$, writing $\theta \succeq \theta'$
        for the corresponding filters.
    \end{enumerate*}
\end{definition}

For example, a car must stop before an obstacle ahead.
Its base filter monitors the stopping margin, the room left if it started braking now, and below a
threshold blends the policy's command with a braking fallback at a fixed ratio.
A harder-braking fallback, a larger threshold, or a blend leaning more on the fallback each stop the
car at least as short, since braking harder or sooner never costs margin; varying them along one
parameter yields a graded family.

\paragraph{Confidence-modulated safety filtering.}
A policy-confidence monitor $\conf : \mathcal{H} \to [0, 1]$ scores how much the task policy can be
trusted on the current information, $\beta_t = \conf(\eta_t)$, with $\beta_t = 1$ meaning full
trust.
A non-increasing schedule $\Gamma : [0, 1] \to \Theta$ then selects a member of a graded family at
every step, $\theta_t = \Gamma(\beta_t)$, so lower confidence selects a more conservative filter,
and the system executes $u_t = \phi_{\theta_t}\big(\eta_t, \task(\eta_t)\big)$.
Because every member satisfies the filter condition, the modulated filter inherits
the base certificate for any schedule as long as the ODD assumptions hold.

\begin{hypothesis}
    \label{hyp:modulation}
    Suppose that (i) the task policy is more likely to act unsafely where its confidence is low, and
    (ii) beyond the ODD, confidence is low where $\monitor$ is not valid, either because those states
    are unfamiliar to the policy or because the model error itself drives the system into
    low-confidence states.
    Then, inside the ODD, a confidence-modulated filter is as performant as its least conservative
    member, $\Gamma(1)$; beyond the ODD, it is as safe as its most conservative member, $\Gamma(0)$.
\end{hypothesis}

Each regime favors an opposite end of the family.
Inside the ODD, every member is certified, so conservativeness buys no safety and only costs
performance: modulation keeps the filter permissive while the policy is confident, and the
conservative members act only where it is not, steering it away from the boundary before abrupt
fallback interventions become necessary~\citep{hu2022sharp}.
Beyond the ODD, nothing is certified, and the margin of the conservative members is what can keep an
unreliable policy safe; modulation applies it where confidence drops, which pays off when that is
where the policy would fail.
One could instead switch between the two ends at the ODD boundary, but in general a deployed system
cannot tell when its filter's assumptions stop holding.
Modulation needs no such knowledge: the same pipeline runs everywhere, guided only by the policy's
confidence.

%% file: sections/05_experiments.tex

\section{Experiments}
\label{sec:experiments}

We evaluate confidence-modulated safety filtering in simulation, using the F1TENTH autonomous racing
benchmark suite of~\texttt{f1tenth\_benchmarks}~\citep{evans2024unifying}.
The task policy, an end-to-end TD3 agent~\citep{fujimoto2018addressing}, maps two stacked LiDAR scans
to steering and speed, and is trained on five tracks with gentle left-hand corners (at most 85$^\circ$).
The confidence monitor $\conf$ maps the reconstruction error of an autoencoder trained on the
policy's own observations~\citep{richter2017safe}.
The filter is a backup CBF~\citep{chen2021backup} with three backup maneuvers,
braking to a stop with the steering held straight, at full left, or at full right; its barrier $h$ is
the clearance to the scanned boundary along the best backup.
At each step, it applies the action closest to the policy's that satisfies the discrete-time CBF
condition $h(\eta_{t+1}) - h(\eta_t) \geq -\gamma h(\eta_t)$~\citep{agrawal2017discrete,xiong2023discrete};
ordered by decreasing $\gamma \in (0, 1]$, these filters form our graded family, and the schedule
$\Gamma(\beta_t)$ sets $\gamma$ from $0.08$ at no trust to $0.55$ at full trust.

The filter is constructed with the simulator's own dynamics at the nominal friction $\mu = 1.05$,
so its assumptions match the simulator there; this defines its ODD, and we go beyond it by lowering
the friction to $\mu = 0.6$ without informing the filter.
We evaluate on three harder tracks with sharp corners (up to 135$^\circ$), including right turns
absent from training.
Unfiltered (Table~\ref{tab:results}), the policy crashes in 45 of 60 laps inside the ODD and in all
laps beyond it.
Inside the ODD, no filter crashes, as its certificate predicts, and modulation closely matches
the permissive filter $\gamma = 0.55$, while being faster than the conservative one at
$\gamma = 0.08$.
Beyond the ODD, the permissive filter crashes in 10 laps, while modulation completes all 60, like the
conservative filter, yet 0.5\,s faster.
These preliminary results are consistent with Hypothesis~\ref{hyp:modulation}.

\begin{figure}[t]
    \centering
    \captionsetup{font=small}
    \includegraphics[width=0.96\linewidth]{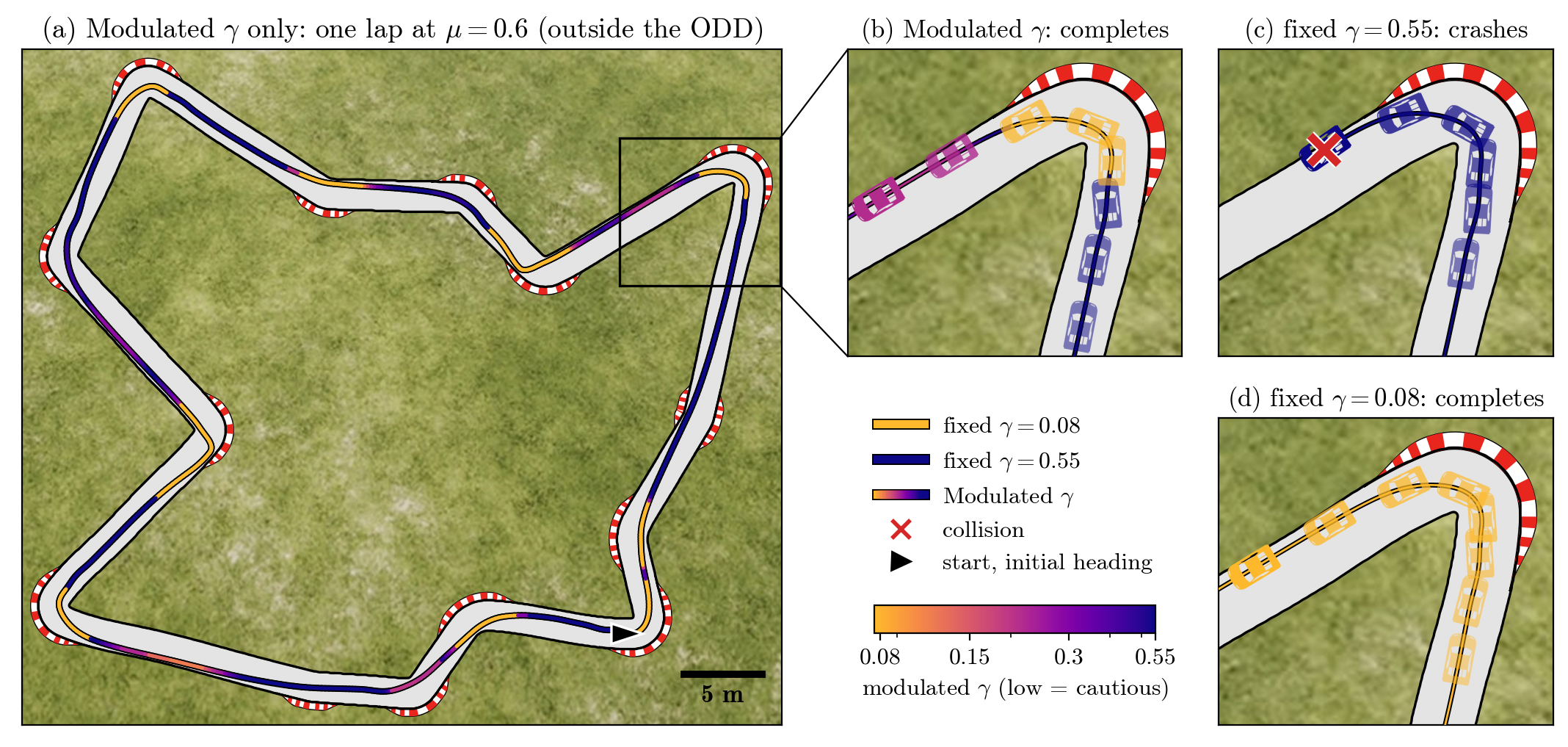}
    \caption{Modulated $\gamma$ outside the ODD. (a) A full lap with the modulated
        filter, colored by $\gamma$: caution (low $\gamma$) is applied at the sharp corners, where the
        policy's confidence is low. (b) Zooms into a sharp corner. (c--d) Same corner for fixed
        $\gamma = 0.55$ (crashes) and fixed $\gamma = 0.08$ (completes).}
    \label{fig:hairpin}
\end{figure}

Fig.~\ref{fig:hairpin} shows one lap beyond the ODD.
The modulated filter stays permissive on straights and gentle corners, which resemble the training
tracks, and turns conservative at the sharp corners, where the policy's confidence drops.
At one such corner, the permissive filter crashes, while the modulated and the conservative filters
complete it.
The autoencoder cannot observe friction: its confidence drops at these corners because their
geometry is unfamiliar, and that is also where low friction invalidates the monitor, the first route
of condition (ii).

\begin{table}[t]
    \centering
    \captionsetup{font=small, position=top}
    \caption{Outcomes for 60 laps per method, inside and outside of the ODD.}
    \label{tab:results}
    \footnotesize
    \setlength{\tabcolsep}{3pt}
    \begin{tabular}{l cccc cccc}
        \toprule
         & \multicolumn{4}{c}{Inside ODD ($\mu = 1.05$)} & \multicolumn{4}{c}{Outside ODD ($\mu = 0.6$)} \\
        \cmidrule(lr){2-5} \cmidrule(lr){6-9}
         & Complete & Timeout & Crash & Lap time (s)
         & Complete & Timeout & Crash & Lap time (s) \\
        \midrule
        No filter & 15 & 0 & 45 & 23.6 $\pm$ 0.1 & 0 & 0 & 60 & -- \\
        Fixed $\gamma = 0.08$ & 60 & 0 & 0 & 31.4 $\pm$ 0.2 & 60 & 0 & 0 & 31.3 $\pm$ 0.2 \\
        Fixed $\gamma = 0.55$ & 60 & 0 & 0 & 30.4 $\pm$ 0.2 & 48 & 2 & 10 & 31.9 $\pm$ 0.3 \\
        Modulated $\gamma$ & 60 & 0 & 0 & 30.2 $\pm$ 0.2 & 60 & 0 & 0 & 30.8 $\pm$ 0.2 \\
        \bottomrule
    \end{tabular}
\end{table}

%% file: sections/06_conclusion.tex

\section{Conclusion}
\label{sec:conclusion}

We proposed modulating a safety filter's conservativeness by the task policy's confidence, with
one pipeline that buys performance inside the filter's ODD and safety margin beyond it.
In preliminary F1TENTH experiments, the modulated filter matched the lap times of the most permissive
fixed filter inside the ODD and the safety of the most conservative one beyond it.
These results remain to be confirmed by broader tests, both in simulation and on real robots.
Since it needs only a confidence score, the pipeline also extends to generalist policies such as robot
foundation models, whose uneven competence across tasks makes a fixed conservativeness especially
costly.